\pdfoutput=1

\documentclass[11pt]{article}

\usepackage{authblk}

\usepackage{EMNLP2023}

\usepackage{times}
\usepackage{latexsym}

\usepackage[T1]{fontenc}

\usepackage[utf8]{inputenc}

\usepackage{graphicx}
\usepackage{caption}
\usepackage{subcaption}
\newcounter{subfig} 
\usepackage{booktabs}

\usepackage{adjustbox}

\usepackage{pifont}
\newcommand{\cmark}{\ding{51}}%
\newcommand{\xmark}{\ding{55}}%

\usepackage{tcolorbox}
\tcbuselibrary{listings,breakable,skins}

\usepackage{microtype}

\usepackage{inconsolata}

\newcommand{\App}{\textsc{SummHelper}}
\newcommand{\AppCompare}{\textsc{OnlySumm}}

\title{\App: Collaborative Human-Computer Summarization}

\newcommand*\samethanks[1][\value{footnote}]{\footnotemark[#1]}

\author[1\Thanks{~~Equal contribution.}]{\bf Aviv Slobodkin}
\author[1\samethanks]{\bf Niv Nachum}
\author[1]{\bf Shmuel Amar}
\author[2\thanks{~~Work done in cooperation with Bar-Ilan University (external and not related to the author’s work at Amazon).}]{\bf Ori Shapira}
\author[1]{\bf Ido Dagan}
{
\makeatletter
\renewcommand\AB@affilsepx{~~~ \protect\Affilfont} \makeatother
\affil[1]{Bar-Ilan University}
\affil[2]{Amazon}
}
\affil[  ]{} 
\affil[  ]{\tt \{lovodkin93, niv252, shmulikamar, obspp18\}@gmail.com}
\affil[  ]{\tt dagan@cs.biu.ac.il}

\begin{document}

\maketitle

\begin{abstract}
Current approaches for text summarization are predominantly automatic, with rather limited space for human intervention and control over the process.
In this paper, we introduce \App{},\footnote{System at \url{https://nlp.biu.ac.il/~sloboda1/SummHelper}, screencast demo at \url{https://www.youtube.com/watch?v=jKzS9RwuccM} and code is available at \url{https://github.com/niv252/SummHelper}} a 2-phase summarization assistant designed to foster human-machine collaboration. 
The initial phase involves content selection, where the system recommends potential content, allowing users to accept, modify, or introduce additional selections. 
The subsequent phase, content consolidation, involves \App{} generating a coherent summary from these selections, which users can then refine using visual mappings between the summary and the source text. 
Small-scale user studies reveal the effectiveness of our application, with participants being especially appreciative of the balance between automated guidance and opportunities for personal input.
\end{abstract}


\begin{figure*}[t!]
    \centering
    \setcounter{subfig}{0} 

    \begin{subfigure}{\linewidth} 
        \centering
        \setlength{\fboxsep}{0pt} 
        \fbox{
            \includegraphics[trim=0 0 1pt 0, clip, width=\linewidth]{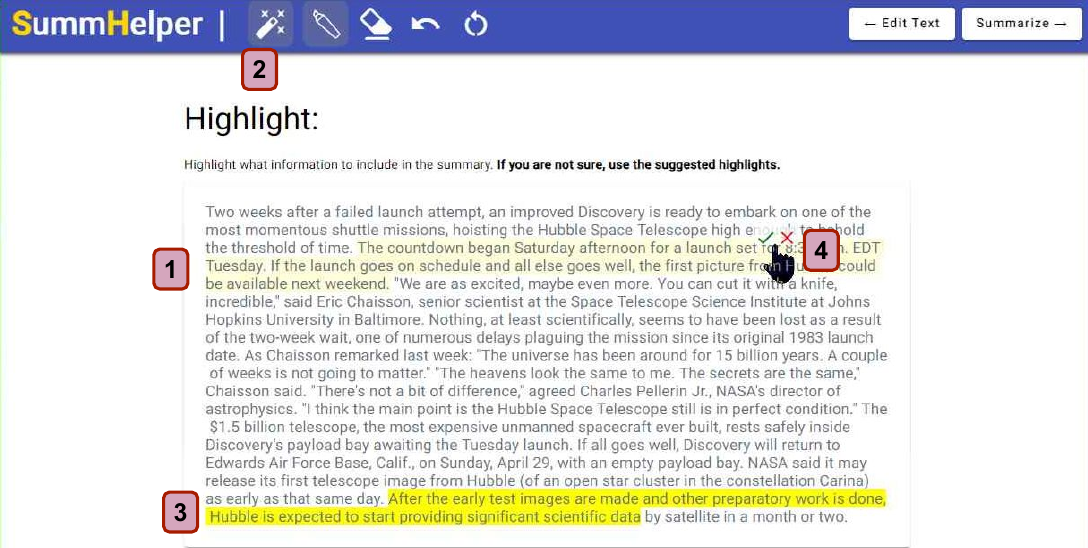}%
        } 
        \stepcounter{subfig} 
        \vspace*{-4mm}
        \caption{(\alph{subfig}) Content selection window}
        \vspace*{4mm}\label{fig:Highlighting_window} 
    \end{subfigure}
    
    \begin{subfigure}{\linewidth} 
        \centering
        \setlength{\fboxsep}{0pt} 
        \fbox{
            \includegraphics[trim=0 0 1pt 0, clip, width=\linewidth]{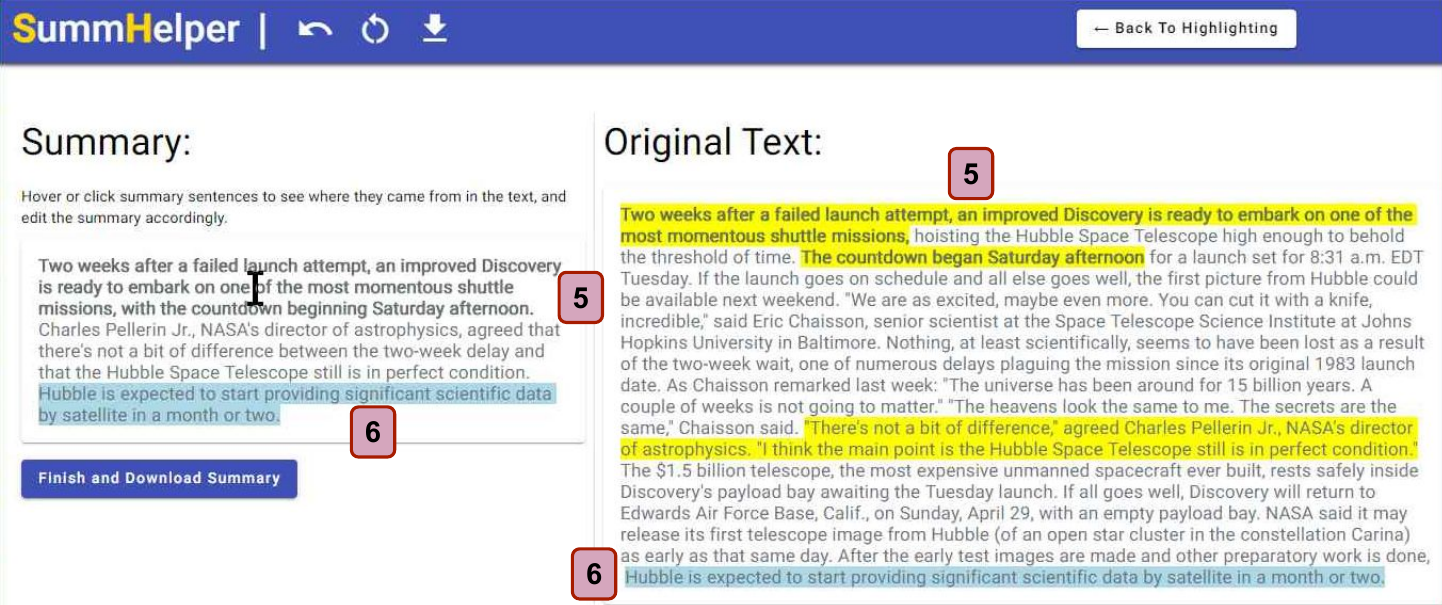}%
        } 
        \stepcounter{subfig} 
        \vspace*{-4mm}
        \caption{(\alph{subfig}) Review and editing window}
        \label{fig:Review_window} 
    \end{subfigure}

    \caption{Our \App{} web application. First, users upload a document and enter the content selection window (\ref{fig:Highlighting_window}) to select what information to include in the summary. Users can receive suggestions from the system (pale yellow; [1]), through the magic wand icon [2]. 
    Any part in the text can be highlighted
    via mouse click-and-drag operations [3]. Users can also accept or reject 
    entire suggested spans
    via the respective \cmark{} and \xmark{} buttons, which appear when hovering over suggestions [4].
    When finishing highlighting, a 
    summary is generated, and users proceed to the reviewing window (\ref{fig:Review_window}), 
    which shows the generated summary and the source text, with highlights, side-by-side.
    Here, hovering over a summary sentence emboldens that sentence and its corresponding aligned source text [5]. Additionally, clicking a summary sentence assigns a persistent blue background to the aligning texts [6].
    Users can edit the summary freely, with alignments updating automatically.} 
    \label{fig:app_printscreens} 
\end{figure*}

\section{Introduction}
\label{sec_introduction}
Text summarization is the task of generating a condensed version of a given text. Most summarization approaches operate in a fully automated pipeline. While efficient, fully automatic summarization does not flexibly enable human intervention and control during the summarization process, which could potentially tune the process to better accommodate user preferences, as well as rectify inevitable mistakes made by models.
Our objective in this paper is to promote such a human-involved approach to summarization, allowing to better tailor the eventual output to real-world user needs, and to synergize the efficiency of the computer with the quality of the human \citep{hoc2000human, pacaux2017designing, flemisch2019joining}.
The process can conveniently support a range of practical scenarios that require individual preferences, such as editors preparing summaries of articles, students condensing notes, or legal practitioners abridging contracts.


To advance such direction, we present \App{}, a 2-stage summarization assistant, which decomposes the summarization pipeline into two natural subtasks---content selection followed by summary generation---and facilitates human-machine cooperation in each subtask. On an input document, the process starts with the selection of content for the summary (\S\ref{subsec:content_selection}). \App{} suggests possible salient content, efficiently pointing users to central information within the text (see [1] in \autoref{fig:Highlighting_window}). Users may accept or reject suggested spans, or highlight any other content to include in the summary ([3] in \autoref{fig:Highlighting_window}).

Upon receiving highlighted content within the text, \App{} subsequently consolidates it and generates a coherent summary (\S\ref{subsec:summary_generation}). This step coincides with the recently introduced Controlled Text Reduction task \citep[CTR;][]{slobodkin-etal-2022-controlled}, which produces a coherent fused version of the content of marked spans (``highlights'') in a source document, as interpreted within the context of the full text.
Once ready, users can review the generated summary and edit any unsatisfactory content. To facilitate inspection, users are presented with a side-by-side display of the summary and the highlighted input (see \autoref{fig:Review_window}), with clearly marked alignments between summary spans and corresponding source text spans ([5] and [6] in \autoref{fig:Review_window}). The automatic alignments aid users in navigating through the input text and identifying summary content that may need editing.

To assess \App{}'s usefulness for generating customized summaries, we conduct two user studies (\S\ref{sec_experiments}), following common human-computer interaction (HCI) methodologies and applying prominent usability questionnaires. These studies indicate the system's utility and user-friendly design for a thorough collaborative summarization process. Notably, users valued the tool's guidance throughout the process, while also appreciating their continuous involvement in refining automatic decisions.

\section{Background and Related Work}
\label{sec_related}
This section provides a brief overview of related lines of work in summarization. These include strategies offering some level of user control (\S\ref{subsec:User-Aware Summarization}), and modular summarization pipelines that separate the task into distinct subtasks (\S\ref{subsec:Modular Summarization}).

\subsection{User Impact on the Summarization Process}\label{subsec:User-Aware Summarization}
Several previous lines of research focused on 
giving users control of the summary content.
In tasks like query-focused \citep{dang-2006-duc, baumel2018qfs} and aspect-based summarization \citep{ahuja-etal-2022-aspectnews, yang-etal-2023-oasum}, the input text is accompanied by a request around which to focus the output summary.
This is a common non-interactive approach for guiding summary content.
Other works adapt the summarization process to specific users by learning their preferences. \citet{hu2012socialPersSumm} and \citet{tepper2018collabot} profile users in order to personalize the summary, via previously discussed aspects in conversations and social connections. Similarly, research on active learning collects summary preferences from users and learns their inclinations toward content and format in order to improve the model's performance \citep{pvs2017feedback, zarinbal2019socialRobot, gao2020preference}.
In these works, user influence is mainly confined to attributes in the input or during model adaptation, leaving the summarization process itself fully automatic. In contrast, our approach supports complete user control and intervention in both content selection and the post-generation phase.

Another line of work focuses on designing interactive tools that provide users with certain means of intervention \textit{during} the summarization process. \citet{yan2011ips} developed a system supporting iterative selection and removal of source sentences in an extractive system summary until a satisfactory summary is obtained. To aid users in making informed decisions, the system helps users track the context in which summary sentences were mentioned in the source texts.
Similarly, \citet{pvs2018sherlock} introduced a tool where users can iteratively select concepts in a system summary to remove from the summary or upon which to further elaborate.
\citet{xie2023interactiveEditing}'s system allows users to edit system summaries by typing text and receiving automatic completion suggestions.
Despite facilitating collaboration with users, these tools start with complete generic system summaries before integrating user feedback. Specifically, they are not well-suited for cases where users wish to include content not present in the initial system summary, or for completely changing its content.
In contrast, our system adapts to user feedback throughout the entire process, allowing users to choose what to include in the summary and assisting them in editing the output to further adjust it to their preferences.

Lastly, interactive exploration systems \citep{shapira2022intsumRL} provide updated summaries for given queries. However, unlike \App{}, such systems aim to allow 
learning about a topic, rather than generating a coherent fine-tuned summary.

\subsection{Modular Summarization}\label{subsec:Modular Summarization}
\App{} is a modular system consisting of separate components, each performing one sub-task, allowing user modifications of that sub-task's output.
Such decomposition has been studied before in the context of fully automated summarization, with several works separating the process into salience detection and generation components \citep{barzilay2005summWithFusion, li2018selectBeforeAbstract, ernst2022procluster}. These works focused on optimizing each component as part of a fully-automatic summarization process in order to improve the overall performance of the model. In contrast, our work uses this modularity to not only improve overall system output, but to also give more control to the user over each step in the summarization process.

\section{The \App{} Application}\label{sec:application}
\App{} is a web application designed for human-computer cooperation in generating human-controlled summaries, shown in \autoref{fig:app_printscreens}.
It consists of two stages: 
(i) computer-assisted content selection via highlighting (\S\ref{subsec:content_selection}), and (ii) automated summary generation according to the selected content followed by machine-assisted reviewing and editing of the generated summary (\S\ref{subsec:summary_generation}).

\subsection{Personalized Content Selection}\label{subsec:content_selection}
The first step focuses on content selection. 
The information to incorporate in the summary is manually selected by highlighting it via mouse click-and-drag operations ([3] in \autoref{fig:Highlighting_window}). Notably, users can also get suggested content from \App{} ([1], pale yellow), by clicking the magic wand icon ([2]).
Users can accept or reject a full suggestion by clicking the \cmark{} and \xmark{} buttons, respectively, which appear when hovering over the suggestion ([4]).

To automatically identify suggested highlights, we deploy the ExtractiveSummarizer model from the TransformerSum library.\footnote{\url{https://transformersum.readthedocs.io/en/latest/}} The model, a RoBERTa\textsubscript{base} \citep{DBLP:journals/corr/abs-1907-11692} trained on the CNN/DailyMail summarization dataset \citep{hermann2015teaching}, operates as a binary classifier. Its function is to assess the significance of each sentence within the text. 
As a subsequent operation, the application selects the $30\%$ highest-ranking sentences
to suggest to the user. 
The choice of this model was influenced by its popularity among extractive summarizers, which are all trained to predict salience. Yet, it can be easily replaced with other content selection models to cater to varying needs.

We note that these recommendations are primarily applicable for \textit{generic} summaries.
The final content selection decision lies with the users, whose judgment and scrutiny of these suggestions, along with the additional selection of non-suggested content, is instrumental in tailoring the summary to their specific preferences.

\subsection{Content Consolidation}\label{subsec:summary_generation}
Once all the desirable content is selected, the next step is to properly consolidate it into a coherent summary. In our setting, \App{} initially auto-generates such a summary, subsequently providing users with guidance for its review and refinement.
For the initial auto-consolidation, we deploy an available Controlled Text Reduction model \citep{slobodkin2023dont}, which is a Flan-T5\textsubscript{large} model \citep{https://doi.org/10.48550/arxiv.2210.11416}, finetuned on the highlights-focused CTR dataset.\footnote{For further details, see Appendix~\ref{appendix_sec_CTR_model}.}
Upon generation, users are presented with the generated summary and the highlighted input text side-by-side (see \autoref{fig:Review_window}). This view facilitates reviewing the summary and editing it when identifying unfavorable outcomes, such as the absence of highlighted content or the inclusion of undesired (non-highlighted or hallucinated) content. To facilitate examination of the summary's compliance with the highlighted content, the user can hover over summary sentences to embolden both the summary sentence and its corresponding alignment in the source text ([5]). 
An alignment can be permanently emphasized with a blue background by clicking on a summary sentence, which remains unaffected when hovering over other sentences ([6]).
To ensure consistent alignment while the summary is being revised by the user, \App{} monitors writing pauses and re-calculates alignments when a pause exceeds one second.

Considering the computational demands of continuous on-the-fly re-alignment, and the alignment feature's primary goal of pointing users to relevant source text sections, we opted for a lexical-matching approach, which is both fast and sufficient for this goal.\footnote{Semantic matching was examined during system development, but was found to have little added value with substantially higher latency.} 
Our approach locates the longest common subsequence (LCS) between the lemmas of each input sentence and each summary sentence, followed by several heuristics to filter out irrelevant LCSs (see \autoref{appendix_sec_alignment_algorithm} for further details).



\section{Experiments and Evaluation}
\label{sec_experiments}
We assess \App{} via two user studies with human subjects, using standard human-computer interaction (HCI) questionnaires.
In the first study, we examine the usability of \App{} for carrying out its purpose, i.e., summarizing an article in a collaborative manner, granting control to the user throughout the process.
The second study compares \App{} to a conventional summarization setup, where a standard auto-generated summary can simply be post-edited without any specialized automated assistance, aiming to assess \App{}'s comparative utility.

\begin{table}[]
\centering
\resizebox{0.85\columnwidth}{!}{%
\begin{tabular}{lcccccc}
\toprule
\textbf{User} & 1 & 2 & 3 & 4 & 5 & 6 \\
\midrule
\textbf{SUS Score} & 95 & 95 & 90 & 67.5 & 90 & 82.5 \\
\bottomrule
\end{tabular}%
}
\caption{SUS scores for each user, calculated based on the ten SUS question scores (see Appendix \ref{appendix_subsec_SUS}).}
\label{tab:SUS}
\end{table}

\subsection{Usability Study}\label{subsec_SUS}
\paragraph{Setup.}
This study aims to gather human feedback regarding the usefulness of \App{} in performing a collaborative, user-guided, summarization process.
Following the discount usability testing principle \citep{nielsen1993usability}, which contends that six evaluators are sufficient for prototype evaluation, we employed six participants for this study.
To simulate a plausible real-world scenario, participants were given the persona of an intern journalist who is required to use the application for writing a summary of a news article.
All participants performed the task twice, over the same two articles, taken from the DUC 2001 dataset,\footnote{\url{https://duc.nist.gov}} in random order.
To assess  \App{}'s helpfulness in different use cases, one article was relatively long, with $\sim$800 tokens, whereas the other contained $\sim$500 tokens.

During the experiments, we observed the users’ activity and employed a ``think aloud'' technique \citep{van1994think} to obtain user remarks.
Upon completing the summaries of both articles, participants filled out the standard System Usability Scale (SUS) questionnaire \citep{brooke1996sus} for subjective usability evaluation, consisting of questions regarding the system's ease of use, ease of learning, and general flow, with an overall score between 0 and 100.

Additionally, after summarizing each article, participants rated the usefulness of various characteristics of the application on a 1 to 5 scale, including the quality of the different models and algorithms used in the system, the intuitiveness of highlighting and unhighlighting content, and the likeliness of them recommending the system.
For more details about the setup, including the full list of the SUS questions and our additional questions, see Appendix~\ref{appendix_subsec_SUS}.

\begin{table}[]
\centering
\resizebox{0.85\columnwidth}{!}{%
\begin{tabular}{lc}
\toprule
\textbf{System Aspect} & \textbf{Score} \\
\midrule
\vspace{1mm} Highlights suggestion model & 3.7 (1.0) \\
\vspace{1mm} Alignments algorithm & \multicolumn{1}{l}{4.3 (1.0)} \\ 
CTR model & \multicolumn{1}{l}{} \\
\hspace{3mm}  {\small Summary coherence} & 4.2 (0.7) \\
\hspace{3mm}  {\small Summary non-redundancy} & 4.6 (0.4) \\
\hspace{3mm}  {\small Highlights coverage} & 4.7 (0.4) \\
\hspace{3mm}  {\small Highlights adherence} & 4.2 (0.7) \\
\hspace{3mm}  {\small Overall satisfaction} & 4.0 (0.7) \\ 
General & \multicolumn{1}{l}{} \\
 \hspace{3mm}  {\small Intuitiveness of highlighting} & \multicolumn{1}{l}{4.5 (0.4)} \\
\hspace{3mm}  {\small Likeliness to recommend } & \multicolumn{1}{l}{4.2 (0.7)} \\
\bottomrule
\end{tabular}%
}
\caption{The average and (StD) results of the Usefulness questionnaire on the 12 sessions (2 articles for 6 participants). See Appendix~\ref{appendix_subsec_SUS} for the full questions.}
\label{tab:usefulness}
\end{table}

\paragraph{Results.}
\autoref{tab:SUS} presents the SUS scores of each of our 6 participants. With the exception of user number 4,\footnote{
This single participant expressed a strong personal preference for a more abstractive automatic summary, even though this is not necessarily a desired goal on its own in our setting.}
the system received scores exceeding 80, thereby affirming the application's ``excellent'' usability \citep{uiux2021susScale}. See \autoref{tab:SUS_avg} in the Appendix for itemized scores.

This favorable trend is further observed in \autoref{tab:usefulness}, which outlines the average ratings on the system features, across the 12 sessions. Overall, users expressed satisfaction with the application, finding \App{}'s features helpful and intuitive, including the initial highlight suggestions 
and the alignment feature.
Furthermore, the generated summaries by the CTR model were viewed as highly satisfactory, and there was a discernible interest among several participants to incorporate such an application into their everyday work (e.g., for summarizing legal contracts as well as prescription drug information).

During the study, we observed that the majority of users felt that the suggested highlights were particularly useful when navigating through the \textit{longer} article as opposed to the \textit{shorter} one. Nevertheless, all users expressed satisfaction with the overall summarization process of \App{} for both articles. They particularly appreciated the two-step procedure encompassing content selection and subsequent review, as it facilitated better text comprehension and instilled greater confidence and control in producing the final output.
Two users expressed a desire for an option to create more abstractive summaries that are less verbatim relative to the highlights. Addressing this feedback, by training more abstractive CTR models or performing a post-hoc abstraction of the generated summary, is an interesting future direction we plan to explore. See Appendix \ref{appendix_subsec_SUS} for more feedback and issues raised by participants.



\subsection{Comparative Usability Test}\label{subsec_USE}

\paragraph{Setup.}
We compared the use of \App{} with a setup that simulates a conventional approach when working with summarization systems. In such setup, the input text is first generically summarized with an automatic summarization model. That summary can then be manually post-edited to meet the specific preferences of the user.
For the summarization model, we used a BART\textsubscript{large} model \citep{DBLP:journals/corr/abs-1910-13461} trained on the CNN/Daily Mail dataset \citep{hermann2015teaching},\footnote{\url{https://huggingface.co/facebook/bart-large-cnn}} selected for its noticeable popularity.
We adapted \App{}'s front-end for this process in order to eliminate a potential influence caused by the application's design. The resulting application comprises two steps: the automatic generation of the generic summary and the review step for manual editing.
During reviewing, users are presented with the input text and the generated summary side-by-side, allowing them to make adaptations to the summary (without the alignment feature).
We refer to this adapted application as \AppCompare{}.

For this experiment, we asked 6 new participants to follow the task described in \S\ref{subsec_SUS}, which involved summarizing a news article, taking the perspective of an intern journalist, once with \App{} on one article, and once with \AppCompare{} on another article (with different orders of articles and applications).
Upon completion of both sessions, participants filled out a questionnaire, adapted from the standard USE Questionnaire \citep{lund2001measuring}.
In the questionnaire, 32 statements are rated on a scale of 1 (\AppCompare{} is preferred) to 5 (\App{} is preferred).
The original 30 USE statements represent 4 dimensions: Usefulness, Ease of Use, Ease of Learning, and Satisfaction (see Appendix~\ref{appendix_subsec_USE} for the full list of statements).
We also added 2 statements to rank users' experience with the key aspects of our summarization process (represented as the fifth dimension in \autoref{tab:USE}). These additional statements were: ``I found it easy to control what information to include in the final summary'' and ``I found it easy to make sure the final summary had all the information I wanted''.
More details elaborating on the study are available in Appendix~\ref{appendix_subsec_USE}.

\begin{table}[]
\centering
\resizebox{0.75\columnwidth}{!}{%
\begin{tabular}{lc}
\toprule
\textbf{Dimension} & \textbf{Score} \\
\midrule
Usefulness & 4.3 (0.5) \\
Ease of Use & 3.6 (0.6) \\
Ease of Learning & 3.1 (0.3) \\
Satisfaction & 4.1 (0.6) \\
Summarization Process & 4.7 (0.5) \\
\bottomrule
\end{tabular}%
}
\caption{The average (StD) results of the five dimensions in the USE questionnaire. A score of 1 represents a preference for \AppCompare{} and 5 prefers \App{}.}
\label{tab:USE}
\end{table}

\paragraph{Results.}
\autoref{tab:USE} presents the scores for each dimension examined, averaged over the corresponding statements and the six participants. Interestingly, despite \App{} consisting of more features and steps than \AppCompare{}, participants did not find it more challenging to learn. Moreover, they reported that \App{} was somewhat more user-friendly. 
\App{} was strongly favored over \AppCompare{} in terms of Usefulness, Satisfaction, and, notably, the Summarization Process, underscoring the practicality of \App{} for preparing customized summaries. 

Importantly, we observed that users tended to be very meticulous when summarizing with \App{}, exhibiting a higher inclination to carefully inspect the text and critically evaluate the inclusion of each piece of information. Indeed, even with the suggested highlights, users cautiously appraised each suggestion and more often selected only sub-segments of it.
In contrast, we found that when summarizing with \AppCompare{}, participants typically skimmed the input text and accepted the generated summaries with minimal adjustments.
Therefore, although using \App{} generally took longer to summarize (11.1 minutes on average, compared to 7.0 minutes with \AppCompare{}), it led to a more thorough summarization process.
This is corroborated by the Usefulness, Satisfaction, and Summarization Process scores in \autoref{tab:USE}, and participants' feedback, which consistently indicated higher confidence and satisfaction with their completed work when using \App{}.


\section{Conclusion}
\label{sec_conclusion}
In this paper, we presented \App{}, a novel summarization assistant, which collaborates with users across two steps: content selection and content consolidation.
The system facilitates user intervention
and supervision along the summarization process, in order to achieve the most suitable output tailored to specific needs.
Preliminary user studies illustrate \App{}'s potential for a thorough and collaborative summarization process, with users expressing satisfaction with the process, as well as the final output.

Future work may include investigating more effective semantic strategies to locate summary-source alignments with acceptable latency.
Additionally, in light of some user feedback, another interesting 
extension includes
developing more abstractive
consolidation and fusion
models, which would offer control over the level of abstractness in the outputs.
Lastly, exploring strategies to scale \App{} to a multi-document setting presents another promising avenue for future investigation.

\section*{Limitations}
\label{sec_limitations}
This demo focuses on the single-document setting. Future work should expand the application's capabilities to the multi-document setting, both in terms of the backend models and in terms of accessibility and intuitiveness of the application's frontend design. Additionally, our tool currently helps users in the reviewing step solely with the alignment functionality. Future work should add additional assistance during this step in the form of suggested improvements to selected unsatisfactory content in the summary, in addition to the alignment feature. 

\section*{Ethics Statement}
\label{sec_ethics}
We conducted the usability (\S\ref{subsec_SUS}) and comparative usability (\S\ref{subsec_USE}) studies in person. Participants volunteered to take part in the study, taking about 40 minutes for the former experiment, or 35 minutes for the latter. A consent form was signed by participants prior to each session, which stressed the fact that the user study was voluntary and that they were encouraged to withdraw if they felt any discomfort. In addition, the form ensured that the participant is at least 18 years of age, and assured that personal details remain anonymous.

The source texts (news articles) used in the user studies were acquired according to the required NIST guidelines (\url{https://duc.nist.gov}).

\section*{Acknowledgements}
This work was supported by the Israel Science Foundation (grant no. 2827/21), and a grant from the Israel Ministry of Science and Technology.
We would also like to thank Hadar Ronen for her guidance in planning the user studies.

\bibliography{main}

\begin{thebibliography}{30}
\expandafter\ifx\csname natexlab\endcsname\relax\def\natexlab#1{#1}\fi

\bibitem[{Ahuja et~al.(2022)Ahuja, Xu, Gupta, Horecka, and
  Durrett}]{ahuja-etal-2022-aspectnews}
Ojas Ahuja, Jiacheng Xu, Akshay Gupta, Kevin Horecka, and Greg Durrett. 2022.
\newblock \href {https://doi.org/10.18653/v1/2022.acl-long.449} {{ASPECTNEWS:
  Aspect-Oriented Summarization of News Documents}}.
\newblock In \emph{Proceedings of the 60th Annual Meeting of the Association
  for Computational Linguistics (Volume 1: Long Papers)}, pages 6494--6506,
  Dublin, Ireland. Association for Computational Linguistics.

\bibitem[{Barzilay and McKeown(2005)}]{barzilay2005summWithFusion}
Regina Barzilay and Kathleen~R. McKeown. 2005.
\newblock \href {https://doi.org/10.1162/089120105774321091} {{Sentence Fusion
  for Multidocument News Summarization}}.
\newblock \emph{Computational Linguistics}, 31(3):297--328.

\bibitem[{Baumel et~al.(2018)Baumel, Eyal, and Elhadad}]{baumel2018qfs}
Tal Baumel, Matan Eyal, and Michael Elhadad. 2018.
\newblock \href {http://arxiv.org/abs/1801.07704} {{Query Focused Abstractive
  Summarization: Incorporating Query Relevance, Multi-Document Coverage, and
  Summary Length Constraints into seq2seq Models}}.

\bibitem[{Brooke(1996)}]{brooke1996sus}
John Brooke. 1996.
\newblock \href
  {https://www.researchgate.net/publication/228593520_SUS_A_quick_and_dirty_usability_scale}
  {{SUS: A Quick and Dirty Usability Scale}}.
\newblock \emph{Usability evaluation in industry}, 189(3):189--194.

\bibitem[{Chung et~al.(2022)Chung, Hou, Longpre, Zoph, Tay, Fedus, Li, Wang,
  Dehghani, Brahma, Webson, Gu, Dai, Suzgun, Chen, Chowdhery, Narang, Mishra,
  Yu, Zhao, Huang, Dai, Yu, Petrov, Chi, Dean, Devlin, Roberts, Zhou, Le, and
  Wei}]{https://doi.org/10.48550/arxiv.2210.11416}
Hyung~Won Chung, Le~Hou, Shayne Longpre, Barret Zoph, Yi~Tay, William Fedus,
  Eric Li, Xuezhi Wang, Mostafa Dehghani, Siddhartha Brahma, Albert Webson,
  Shixiang~Shane Gu, Zhuyun Dai, Mirac Suzgun, Xinyun Chen, Aakanksha
  Chowdhery, Sharan Narang, Gaurav Mishra, Adams Yu, Vincent Zhao, Yanping
  Huang, Andrew Dai, Hongkun Yu, Slav Petrov, Ed~H. Chi, Jeff Dean, Jacob
  Devlin, Adam Roberts, Denny Zhou, Quoc~V. Le, and Jason Wei. 2022.
\newblock \href {https://doi.org/10.48550/ARXIV.2210.11416} {{Scaling
  Instruction-Finetuned Language Models}}.

\bibitem[{Dang(2006)}]{dang-2006-duc}
Hoa~Trang Dang. 2006.
\newblock \href {https://aclanthology.org/W06-0707} {{DUC 2005: Evaluation of
  Question-Focused Summarization Systems}}.
\newblock In \emph{Proceedings of the Workshop on Task-Focused Summarization
  and Question Answering}, pages 48--55, Sydney, Australia. Association for
  Computational Linguistics.

\bibitem[{Ernst et~al.(2022)Ernst, Caciularu, Shapira, Pasunuru, Bansal,
  Goldberger, and Dagan}]{ernst2022procluster}
Ori Ernst, Avi Caciularu, Ori Shapira, Ramakanth Pasunuru, Mohit Bansal, Jacob
  Goldberger, and Ido Dagan. 2022.
\newblock \href {https://doi.org/10.18653/v1/2022.naacl-main.128}
  {{Proposition-Level Clustering for Multi-Document Summarization}}.
\newblock In \emph{Proceedings of the 2022 Conference of the North American
  Chapter of the Association for Computational Linguistics: Human Language
  Technologies}, pages 1765--1779, Seattle, United States. Association for
  Computational Linguistics.

\bibitem[{Flemisch et~al.(2019)Flemisch, Abbink, Itoh, Pacaux-Lemoine, and
  Wessel}]{flemisch2019joining}
Frank Flemisch, David~A Abbink, Makoto Itoh, M-P Pacaux-Lemoine, and Gina
  Wessel. 2019.
\newblock \href
  {https://www.researchgate.net/publication/335076750_Joining_the_blunt_and_the_pointy_end_of_the_spear_towards_a_common_framework_of_joint_action_human-machine_cooperation_cooperative_guidance_and_control_shared_traded_and_supervisory_control}
  {{Joining the blunt and the pointy end of the spear: towards a common
  framework of joint action, human--machine cooperation, cooperative guidance
  and control, shared, traded and supervisory control}}.
\newblock \emph{Cognition, Technology \& Work}, 21:555--568.

\bibitem[{Gao et~al.(2020)Gao, Meyer, and Gurevych}]{gao2020preference}
Yang Gao, Christian~M. Meyer, and Iryna Gurevych. 2020.
\newblock \href {https://doi.org/10.1007/s10791-019-09367-8} {{Preference-Based
  Interactive Multi-Document Summarisation}}.
\newblock \emph{Information Retrieval}, 23(6):555–585.

\bibitem[{Hermann et~al.(2015)Hermann, Kočiský, Grefenstette, Espeholt, Kay,
  Suleyman, and Blunsom}]{hermann2015teaching}
Karl~Moritz Hermann, Tomáš Kočiský, Edward Grefenstette, Lasse Espeholt,
  Will Kay, Mustafa Suleyman, and Phil Blunsom. 2015.
\newblock \href {http://arxiv.org/abs/1506.03340} {{Teaching Machines to Read
  and Comprehend}}.

\bibitem[{Hoc(2000)}]{hoc2000human}
Jean-Michel Hoc. 2000.
\newblock \href {https://pubmed.ncbi.nlm.nih.gov/10929820/} {{From
  human--machine interaction to human--machine cooperation}}.
\newblock \emph{Ergonomics}, 43(7):833--843.

\bibitem[{Hu et~al.(2012)Hu, Ji, Teng, and Guo}]{hu2012socialPersSumm}
Po~Hu, Donghong Ji, Chong Teng, and Yujing Guo. 2012.
\newblock \href {https://aclanthology.org/C12-1075} {{Context-Enhanced
  Personalized Social Summarization}}.
\newblock In \emph{Proceedings of {COLING} 2012}, pages 1223--1238, Mumbai,
  India. The COLING 2012 Organizing Committee.

\bibitem[{Lewis et~al.(2019)Lewis, Liu, Goyal, Ghazvininejad, Mohamed, Levy,
  Stoyanov, and Zettlemoyer}]{DBLP:journals/corr/abs-1910-13461}
Mike Lewis, Yinhan Liu, Naman Goyal, Marjan Ghazvininejad, Abdelrahman Mohamed,
  Omer Levy, Veselin Stoyanov, and Luke Zettlemoyer. 2019.
\newblock \href {http://arxiv.org/abs/1910.13461} {{BART: Denoising
  Sequence-to-Sequence Pre-training for Natural Language Generation,
  Translation, and Comprehension}}.
\newblock \emph{CoRR}, abs/1910.13461.

\bibitem[{Li et~al.(2018)Li, Xiao, Lyu, and Wang}]{li2018selectBeforeAbstract}
Wei Li, Xinyan Xiao, Yajuan Lyu, and Yuanzhuo Wang. 2018.
\newblock \href {https://doi.org/10.18653/v1/D18-1205} {{Improving Neural
  Abstractive Document Summarization with Explicit Information Selection
  Modeling}}.
\newblock In \emph{Proceedings of the 2018 Conference on Empirical Methods in
  Natural Language Processing}, pages 1787--1796, Brussels, Belgium.
  Association for Computational Linguistics.

\bibitem[{Liu et~al.(2019)Liu, Ott, Goyal, Du, Joshi, Chen, Levy, Lewis,
  Zettlemoyer, and Stoyanov}]{DBLP:journals/corr/abs-1907-11692}
Yinhan Liu, Myle Ott, Naman Goyal, Jingfei Du, Mandar Joshi, Danqi Chen, Omer
  Levy, Mike Lewis, Luke Zettlemoyer, and Veselin Stoyanov. 2019.
\newblock \href {http://arxiv.org/abs/1907.11692} {{RoBERTa: A Robustly
  Optimized BERT Pretraining Approach}}.
\newblock \emph{CoRR}, abs/1907.11692.

\bibitem[{Lund(2001)}]{lund2001measuring}
Arnold~M Lund. 2001.
\newblock \href
  {https://www.researchgate.net/publication/230786746_Measuring_Usability_with_the_USE_Questionnaire}
  {{Measuring Usability with the USE Questionnaire}}.
\newblock \emph{Usability interface}, 8(2):3--6.

\bibitem[{Nielsen(1993)}]{nielsen1993usability}
Jakob Nielsen. 1993.
\newblock \href {https://dl.acm.org/doi/book/10.5555/2821575} {{Usability
  Engineering}}.

\bibitem[{Pacaux-Lemoine et~al.(2017)Pacaux-Lemoine, Trentesaux, Rey, and
  Millot}]{pacaux2017designing}
Marie-Pierre Pacaux-Lemoine, Damien Trentesaux, Gabriel~Zambrano Rey, and
  Patrick Millot. 2017.
\newblock \href
  {https://www.sciencedirect.com/science/article/pii/S0360835217302188}
  {{Designing intelligent manufacturing systems through Human-Machine
  Cooperation principles: A human-centered approach}}.
\newblock \emph{Computers \& Industrial Engineering}, 111:581--595.

\bibitem[{P.V.S. et~al.(2018)P.V.S., H\"{a}ttasch, \"{O}zyurt, Binnig, and
  Meyer}]{pvs2018sherlock}
Avinesh P.V.S., Benjamin H\"{a}ttasch, Orkan \"{O}zyurt, Carsten Binnig, and
  Christian~M. Meyer. 2018.
\newblock \href {https://doi.org/10.14778/3229863.3236220} {{Sherlock: A System
  for Interactive Summarization of Large Text Collections}}.
\newblock \emph{Proc. VLDB Endow.}, 11(12):1902–1905.

\bibitem[{P.V.S and Meyer(2017)}]{pvs2017feedback}
Avinesh P.V.S and Christian~M. Meyer. 2017.
\newblock \href {https://doi.org/10.18653/v1/P17-1124} {{Joint Optimization of
  User-desired Content in Multi-document Summaries by Learning from User
  Feedback}}.
\newblock In \emph{Proceedings of the 55th Annual Meeting of the Association
  for Computational Linguistics (Volume 1: Long Papers)}, pages 1353--1363,
  Vancouver, Canada. Association for Computational Linguistics.

\bibitem[{Shapira et~al.(2022)Shapira, Pasunuru, Bansal, Dagan, and
  Amsterdamer}]{shapira2022intsumRL}
Ori Shapira, Ramakanth Pasunuru, Mohit Bansal, Ido Dagan, and Yael Amsterdamer.
  2022.
\newblock \href {https://doi.org/10.18653/v1/2022.naacl-main.184} {{Interactive
  Query-Assisted Summarization via Deep Reinforcement Learning}}.
\newblock In \emph{Proceedings of the 2022 Conference of the North American
  Chapter of the Association for Computational Linguistics: Human Language
  Technologies}, pages 2551--2568, Seattle, United States. Association for
  Computational Linguistics.

\bibitem[{Slobodkin et~al.(2023)Slobodkin, Caciularu, Hirsch, and
  Dagan}]{slobodkin2023dont}
Aviv Slobodkin, Avi Caciularu, Eran Hirsch, and Ido Dagan. 2023.
\newblock \href {http://arxiv.org/abs/2310.09017} {Dont add, dont miss:
  Effective content preserving generation from pre-selected text spans}.

\bibitem[{Slobodkin et~al.(2022)Slobodkin, Roit, Hirsch, Ernst, and
  Dagan}]{slobodkin-etal-2022-controlled}
Aviv Slobodkin, Paul Roit, Eran Hirsch, Ori Ernst, and Ido Dagan. 2022.
\newblock \href {https://aclanthology.org/2022.emnlp-main.385} {{Controlled
  Text Reduction}}.
\newblock In \emph{Proceedings of the 2022 Conference on Empirical Methods in
  Natural Language Processing}, pages 5699--5715, Abu Dhabi, United Arab
  Emirates. Association for Computational Linguistics.

\bibitem[{Tepper et~al.(2018)Tepper, Hashavit, Barnea, Ronen, and
  Leiba}]{tepper2018collabot}
Naama Tepper, Anat Hashavit, Maya Barnea, Inbal Ronen, and Lior Leiba. 2018.
\newblock \href {https://doi.org/10.1145/3159652.3160588} {{Collabot:
  Personalized Group Chat Summarization}}.
\newblock In \emph{Proceedings of the Eleventh ACM International Conference on
  Web Search and Data Mining}, WSDM '18, page 771–774, New York, NY, USA.
  Association for Computing Machinery.

\bibitem[{UIUX-Trend(2021)}]{uiux2021susScale}
UIUX-Trend. 2021.
\newblock {Measuring and Interpreting System Usability Scale - UIUX Trend}.
\newblock \url{https://uiuxtrend.com/measuring-system-usability-scale-sus/}.
\newblock Accessed: 2023-08-01.

\bibitem[{Van~Someren et~al.(1994)Van~Someren, Barnard, and
  Sandberg}]{van1994think}
Maarten Van~Someren, Yvonne~F Barnard, and J~Sandberg. 1994.
\newblock \href
  {https://www.researchgate.net/publication/215439100_The_Think_Aloud_Method_-_A_Practical_Guide_to_Modelling_CognitiveProcesses}
  {{The Think Aloud Method: A Practical Approach to Modelling Cognitive
  Processes}}.
\newblock \emph{London: AcademicPress}, 11:29--41.

\bibitem[{Xie et~al.(2023)Xie, Wang, Chen, Xiong, and
  He}]{xie2023interactiveEditing}
Yujia Xie, Xun Wang, Si-Qing Chen, Wayne Xiong, and Pengcheng He. 2023.
\newblock \href {http://arxiv.org/abs/2306.03067} {{Interactive Editing for
  Text Summarization}}.

\bibitem[{Yan et~al.(2011)Yan, Nie, and Li}]{yan2011ips}
Rui Yan, Jian-Yun Nie, and Xiaoming Li. 2011.
\newblock \href {https://aclanthology.org/D11-1124} {{Summarize What You Are
  Interested In: An Optimization Framework for Interactive Personalized
  Summarization}}.
\newblock In \emph{Proceedings of the 2011 Conference on Empirical Methods in
  Natural Language Processing}, pages 1342--1351, Edinburgh, Scotland, UK.
  Association for Computational Linguistics.

\bibitem[{Yang et~al.(2023)Yang, Song, Cho, Wang, Pan, Petzold, and
  Yu}]{yang-etal-2023-oasum}
Xianjun Yang, Kaiqiang Song, Sangwoo Cho, Xiaoyang Wang, Xiaoman Pan, Linda
  Petzold, and Dong Yu. 2023.
\newblock \href {https://aclanthology.org/2023.findings-acl.268} {{OASum:
  Large-Scale Open Domain Aspect-based Summarization}}.
\newblock In \emph{Findings of the Association for Computational Linguistics:
  ACL 2023}, pages 4381--4401, Toronto, Canada. Association for Computational
  Linguistics.

\bibitem[{Zarinbal et~al.(2019)Zarinbal, Mohebi, Mosalli, Haratinik,
  Jabalameli, and Bayatmakou}]{zarinbal2019socialRobot}
Marzieh Zarinbal, Azadeh Mohebi, Hesamoddin Mosalli, Razieh Haratinik, Zahra
  Jabalameli, and Farnoush Bayatmakou. 2019.
\newblock \href
  {https://link.springer.com/chapter/10.1007/978-3-030-26118-4_32} {{A New
  Social Robot for Interactive Query-Based Summarization: Scientific Document
  Summarization}}.
\newblock In \emph{Interactive Collaborative Robotics}, pages 330--340, Cham.
  Springer International Publishing.

\end{thebibliography}
\bibliographystyle{acl_natbib}

\appendix
\section{Alignment Algorithm}\label{appendix_sec_alignment_algorithm}
In the reviewing phase, the system aids users in 
comparing the highlighted input text to the generated output summary, in order to spot any potential disapprovals in the summary.
This is achieved by automatically identifying text from the input that aligns with each sentence in the summary, and clearly marking it (\S\ref{subsec:summary_generation}) for the user.
To find these alignments, the system first performs sentence tokenization on the input source text and the generated summary.
For each pair of summary and input sentences, it then calculates the longest common subsequence (LCS) of their lemmas.

\begin{figure}[t!]
\centering
    \includegraphics[width=0.9\columnwidth]{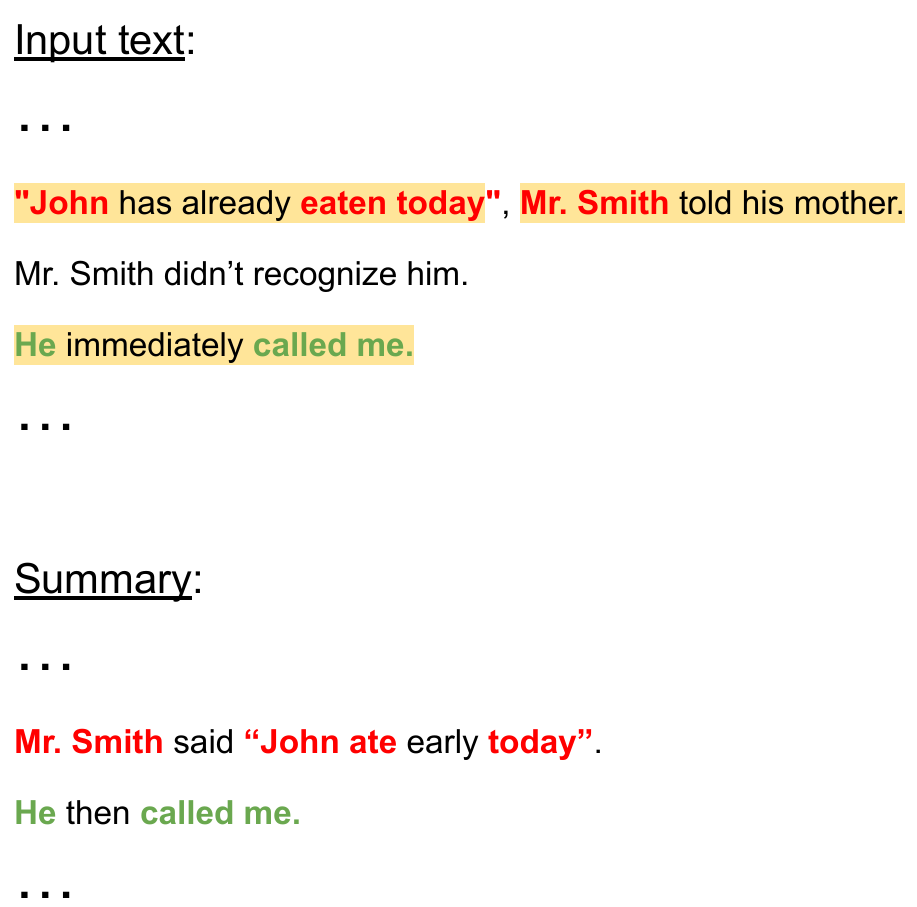}
    \caption{An example of the alignment algorithm for an extract of the highlighted input text and that of the respective summary. The first lemma-based LCS between the first sentences of the summary and input is \textit{``John eat today''} (bold and red), which has $\geq3$ content words (John, eat, today) and is thus retained. The second LCS, \textit{``Mr. Smith''}, contains $\geq25\%$ of the second highlighted span's (\textit{``Mr.''}, \textit{``Smith''}, \textit{``tell''}, \textit{``mother''}) content words, and is also retained. On the other hand, the LCS \textit{``Mr. Smith''} between the first summary sentence and the second input sentence, having only two content words and lacking overlap with any highlighted span, is filtered out.  For the second summary sentence and the third input sentence, the only LCS, \textit{``He called me''} (bold and green), comprises a single content word (\textit{``called''}) which covers $100\%$ of the third highlight's content words and is thereby retained.}
    \label{fig:alignment_algorithm_example}
\end{figure}

To filter out insignificant alignments, LCSs containing less than three content tokens (neither stop words nor punctuation), denoted \textit{short LCS}s, are disregarded.
For instance, as demonstrated in \autoref{fig:alignment_algorithm_example}, the LCS ``John eat today'' between the first sentences of the summary and input consists of three content words and is thus preserved.
In contrast, the LCS ``Mr. Smith'' between the first summary sentence and the second input sentence, having only two content words, is discarded.
For alignment \textit{within highlights}, a short LCS is still retained if it covers at least $25\%$ of the highlighted span's content lemmas.
For instance, even though the LCS ``he call me'' of the last sentences of the summary and input in \autoref{fig:alignment_algorithm_example} contains only one content lemma (``call''), it covers $100\%$ of the highlight's content lemmas and is thus retained.

Finally, the alignment algorithm also addresses cases where the CTR model reorders content within input sentences. An LCS procedure is not well-suited for such situations. To this end, the algorithm iteratively calculates four LCSs for each pair of summary and input sentences.
After each iteration, the part of the summary sentence contributing to the LCS is omitted, enabling shorter LCSs to be identified. 
For example, after identifying the LCS of the first sentences of the input and summary in \autoref{fig:alignment_algorithm_example} (``John eat today''), the algorithm generates a variant of the summary sentence by excluding the LCS, resulting in ``Mr. Smith said early''.
It then identifies the LCS ``Mr. Smith'' between this variant and the first input sentence, which is preserved as it covers $50\%$ of the second highlighted span's content lemmas (``Mr.'', ``Smith'', ``tell'', ``mother'').

\section{CTR Model}\label{appendix_sec_CTR_model}
Controlled Text Reduction \citep[CTR;][]{slobodkin-etal-2022-controlled}, is a recently introduced task, which takes as input a text with pre-selected marked spans (``highlights'') and expects a coherent version of the text, covering exactly the content of these highlights. It handles coherence issues relating to discourse and coreference.
This task conforms with our summary generation process, and we hence employ an available Controlled Text Reduction model.\footnote{\url{https://github.com/lovodkin93/CTR_instruction_finetuning}} 
This model is a a Flan-T5\textsubscript{large} model \citep{https://doi.org/10.48550/arxiv.2210.11416}, finetuned on the highlights-focused CTR dataset.
Following \citet{slobodkin-etal-2022-controlled}, highlights are incorporated into the input text with special markups, \textit{<extra\_id\_1>} and \textit{<extra\_id\_2>}, marking the beginning and end of each highlighted span, respectively. 
In our configuration, we set the maximum input length to 4096 and the maximum target length to 400. A greedy decoding strategy was used in order to optimize the decoding speed. Other parameters are kept consistent with the pre-defined generation parameters of the model.

\section{Experimental Details}\label{appendix_sec_experimental_details}
In this work, we performed a usability study and a system comparison experiment (\S\ref{sec_experiments}) to assess the utility of our application.

\begin{figure}[t]

\lstdefinestyle{promptStyle}
{
    basicstyle={\footnotesize\ttfamily}, 
    numbers=none, 
    xrightmargin=0.5em,
    showstringspaces=false,
    showspaces=false,
    showtabs=false,
    tabsize=2,
    breaklines=true,
    flexiblecolumns=true,
    escapeinside={<@}{@>},
    breakatwhitespace=true
}

\newtcblisting{mylisting}[1]{
  enhanced,
  listing only,
  boxrule=0.8pt,
  sharp corners=downhill,
  top=0mm,
  bottom=0mm,
  left=2mm,
  right=0mm,
  boxsep=0mm,
  colframe=black,
  colback=white,
  listing options={
    style=#1
  }
}

\begin{mylisting}{promptStyle}
As an intern reporter, your assignment is to 
study two articles written by a senior 
journalist, and write a summary for each 
article, suitable for sharing on social media 
platforms.
This task forms a critical part of your 
internship evaluation, hence meticulous 
attention to detail is mandatory.
You're granted access to an application that 
can assist you in accomplishing this task. 
However, it's crucial that the final summary 
remains a testament to your individual effort 
and understanding of the articles.
\end{mylisting}

\caption{The instructions given to the user study participants.}
\label{fig:user_study_assignment}
\end{figure}

\subsection{System Usability Tests}\label{appendix_subsec_SUS}
For the usability study, six participants were gathered based on previous acquaintance. These participants varied in their age (28-33), gender, and occupation. 
Each session took approximately 40 minutes. A participant started by filling out an experiment participation consent form. Next, the different elements of the application were explained and demonstrated to the participant. Then, the participant was asked to experiment with the application on an example article, to reduce the learning curve of using the system for the first time.
Once this onboarding stage was over, the experimentee was presented with the assignment (see \autoref{fig:user_study_assignment}). The participants conducted the experiments on two articles, one with $\sim$800 tokens and another with  $\sim$500 tokens, in a random order.

\paragraph{SUS questionnaire.} The System Usability Scale (SUS) questionnaire \citep{brooke1996sus} was filled out once by each participant after completing both article summaries, with the following 10 questions being rated on a scale from 1 (``strongly disagree'') to 5 (``strongly agree''):

\begin{enumerate}
\item{I think that I would like to use this system frequently.}

\item{I found the system unnecessarily complex.}

\item{I thought the system was easy to use.}

\item{I think that I would need the support of a technical person to be able to use this system.}

\item{I found the various functions in this system were well integrated.}

\item{I thought there was too much inconsistency in this system.}

\item{I would imagine that most people would learn to use this system very quickly.}

\item{I found the system very cumbersome to use.}

\item{I felt very confident using the system.}

\item{I needed to learn a lot of things before I could get going with this system.}

\end{enumerate}

The SUS scores (\autoref{tab:SUS}) were calculated using the procedure by \citet{brooke1996sus}, as follows.
Initially, the score contributions from each item were summed up, with each item's contribution ranging in a 0 to 4 scale. For the odd-numbered items (1,3,5,7, and 9), the score contribution was determined as the scale position minus 1. Conversely, for the even-numbered items (2,4,6,8, and 10), the contribution was calculated as 5 minus the scale position. 
This sum was then multiplied by 2.5 to compute the overall SUS value for each user, with scores having a range of 0 to 100.
We also calculated the average (StD) score for each question, as delineated in \autoref{tab:SUS_avg}. 

\begin{table}[]
\centering
\resizebox{\columnwidth}{!}{%
\begin{tabular}{lc}
\toprule
\textbf{SUS Question} & \textbf{Average Score} \\ \midrule
\begin{tabular}[c]{@{}l@{}}I think that I would like to use this system\\ frequently.\end{tabular} & 4.17 (0.98) \\ \midrule
I found the system unnecessarily complex. & 1.83 (0.98) \\ \midrule
I thought the system was easy to use. & 4.33 (0.52) \\ \midrule
\begin{tabular}[c]{@{}l@{}}I think that I would need the support of a \\ technical person to be able to use this system.\end{tabular} & 1.33 (0.82) \\  \midrule
\begin{tabular}[c]{@{}l@{}}I found the various functions in this system \\ were well integrated.\end{tabular} & 4.17 (0.41) \\  \midrule
\begin{tabular}[c]{@{}l@{}}I thought there was too much inconsistency \\ in this system.\end{tabular} & 1.17 (0.41) \\  \midrule
\begin{tabular}[c]{@{}l@{}}I would imagine that most people would learn \\ to use this system very quickly.\end{tabular} & 4.67 (0.52) \\  \midrule
I found the system very cumbersome to use. & 1.33 (0.52) \\ \midrule
I felt very confident using the system. & 4.50 (0.84) \\ \midrule
\begin{tabular}[c]{@{}l@{}}I needed to learn a lot of things before I could \\ get going with this system.\end{tabular} & 1.50 (0.84) \\
\bottomrule
\end{tabular}%
}
\caption{The average (StD) score of the ten SUS questions asked after the usability study, on a scale of 1 to 5.}
\label{tab:SUS_avg}
\end{table}

\paragraph{Usefulness questionnaire.}
After summarizing each of the two articles, the users filled out a usefulness questionnaire (see results in \autoref{tab:usefulness}), where they
were asked to rate the following 9 questions on a scale of 1 (``strongly disagree'') to 5 (``strongly agree''), addressing the different components in our system:

\begin{enumerate}
\item{For the requirements of the given task, the initial highlights were very helpful.}

\item{The alignments were helpful in assessing the content of the final summary.}

\item{It was intuitive to highlight and unhighlight information.}

\item{I would recommend this app for another intern journalist in my company.}

\end{enumerate} 
~~Overall, the summary output by the system was:

\begin{enumerate}
\setcounter{enumi}{4}
\item{Coherent}
\item{Non-Redundant}
\item{Highlights were covered fully}
\item{Did not cover unhighlighted content}
\item{To my satisfaction}
\end{enumerate}

\paragraph{Comments raised by participants.}
During the sessions, we collected comments and ideas for improvements raised by the participants. All the participants were very impressed with the summaries generated by the CTR model. Additionally, several users expressed their satisfaction with the modular process, stating that their continuous involvement was crucial for achieving the optimal summary. Users especially appreciated the side-by-side presentation of the highlighted input text and the summary, combined with the alignment feature, which helped them to both stay connected to the source text and optimize their navigation through it.
For improvements, one suggestion was to enable generation of more abstractive summaries, that do not align as much with the highlights' phrasing. Additional suggestions included making a different icon for exiting erase mode and entering highlight mode in the content selection window,\footnote{In the first system version, there was only an icon to enter erasing mode, and in order to exit the erasing mode and enter highlighting mode, users needed to click this icon again.} enabling a dynamic number of suggestions proportionate to the text's length,\footnote{In the first system version, there were always 3 suggestions.} and enabling the option to go back to the beginning of the process by clicking the application's name in the toolbar.

\subsection{System Comparison Experiment}\label{appendix_subsec_USE}
For the comparative experiment, we gathered 6 new participants, also based on previous acquaintance. These participants varied in their age (24-35), gender, and occupation. 
Each session took approximately 35 minutes, which started with a participant filling out the same participation form as in the system usability tests (see Appendix~\ref{appendix_subsec_SUS}).
Similarly to the usability test setting, prior to the actual experiment, the different elements of each of the two applications were explained and demonstrated to the participant, and they were asked to experiment with the system on an article.
Once the participant felt confident with their understanding of each application, they were presented with the assignment in \autoref{fig:user_study_assignment} and asked to complete it on the same 2 articles as in the system usability tests, once with \App{} and once with \AppCompare{} (in different orders and different article-model pairings).

\paragraph{Questionnaire.} After completing both articles, the participant answered a comparative usability questionnaire, adapted from the standard USE Questionnaire \citep{lund2001measuring}, as mentioned in \S\ref{subsec_USE}. The original questionnaire consists of 30 statements, divided into 4 dimensions: Usefulness, Ease of Use, Ease of Learning, and Satisfaction. These questions are:

\begin{itemize}

\item \underline{Usefulness} 
\begin{enumerate}
 \item{It helps me be more effective.}

 \item{It helps me be more productive.}

 \item{It is useful.}

 \item{It gives me more control over output.}

 \item{It makes it easier to achieve the desired output.}

 \item{It saves me time when I use it.}

 \item{It meets my needs in addressing the task.}

 \item{It does everything I would expect it to do.}

\end{enumerate}

\item \underline{Ease of Use}

\begin{enumerate}
\setcounter{enumi}{8}

 \item{It is easy to use.}

 \item{It is simple to use.}

 \item{It is user-friendly.}

 \item{It requires the fewest steps possible to accomplish the task.}

 \item{It is flexible.}

 \item{Using it is effortless.}

 \item{I can use it without written instructions.}

 \item{I don't notice any inconsistencies as I use it.}

 \item{Both occasional and regular users would like it.}

 \item{I can recover from mistakes quickly and easily.}

 \item{I can use it successfully every time.}

\end{enumerate} 

\item \underline{Ease of Learning}

\begin{enumerate}
\setcounter{enumi}{19}

 \item{I learned to use it quickly.}

 \item{I easily remember how to use it.}

 \item{It is easy to learn to use it.}

 \item{I quickly became skillful with it.}

\end{enumerate}

\item \underline{Satisfaction}

\begin{enumerate}
\setcounter{enumi}{23}

 \item{I am satisfied with it.}

 \item{I would recommend it to a friend.}

 \item{It is fun to use.}

 \item{It works the way I want it to work.}

 \item{It is wonderful.}

 \item{I feel I need to have it.}

 \item{It is pleasant to use.}

\end{enumerate}

\end{itemize}

For each statement, participants were asked to rate it on a scale from 1 (preferred \AppCompare{}) to 5 (preferred \App{}). In addition to those statements, we added two more statements, in order to rate the participants' experience with the key aspects of the \underline{Summarization Process}:

\begin{enumerate}
    \setcounter{enumi}{30}
    \item I found it easy to control what information to include in the final summary.
    \item I found it easy to make sure the final summary had all the information I wanted.
\end{enumerate}

\paragraph{Observations and general feedback.}
Overall, all participants favored \App{} over \AppCompare{}. They especially appreciated the alignment feature, with one participant who started with \App{}, and expressed frustration with the absence of the alignment feature in \AppCompare{}. Additionally, we observed that all 6 users were meticulous when working with \App{}, and appraised each suggestion very carefully, as well as non-suggested content. Alternatively, when working with \AppCompare{}, 4 out of the 6 participants simply skimmed the article and were quick to accept the generated summary with minimal adjustments. This shows \App{}'s potential to foster a more thorough and productive summarization process.

\end{document}